# Natural Language Understanding Based on Semantic Relations between Sentences

Hyeok Kong
Faculty of mathematics, Kim Il Sung University, D.P.R.K

**Abstract:** In this paper, we define event expression over sentences of natural language and semantic relations between events. Based on this definition, we formally consider text understanding process having events as basic unit.

## 1. Event and Semantic Relation

We introduce semantic representation model of sentences for text understanding, which is based on event expression by considering a sentence as a description of event. We first give event expression consisting of certain items which is called case-relations, where event expression represents the meaning of a sentence.

Formally, an event corresponding to a sentence S can be represented as follows:

$$event(e_1,\cdots,e_n) \quad \text{or} \quad event(s)$$

In some cases, we simply write *event* as *e* and omit *event*. Let $e_i (i=1,\cdots,n)$ be a tuple $(A_i, v(A_i))$ or $(A_i, T_i)$. Here, $A_i$ is a case-relation shown in Table 1 and $v(A_i)$ is a value of case-relation $A_i$ which takes a Korean word or a variable. We call a case-relation taking a variable case-relation with undefined value.

$T_i$ is an event expression which is the value of case-relation $A_i$. This reflects the internal nesting of natural language sentences and corresponds to subordinate clause of principal clause of a sentence. We introduced 14 case-relations in domain of our text understanding system(Table 1).

There exists semantic relation between sentences of a text. Naturally, there exists semantic relation between events corresponding to these sentences.

Table 1．The meaning of case-relation

| case-relation | meaning |
|---|---|
| *actor* | Actor of event |
| *action* | Action of event |
| *verb2* | Description of state or location of event |
| *isa* | Judgment predicate |
| *time* | Time of action |
| *loc* | Location of action |
| *way* | Way of action |
| *obj* | Object of action |
| *source* | Source of action |
| *to* | Goal of action |
| *det* | Determination of person or object |
| *mod* | Mode of action |
| *number* | Number of objects |
| *no* | Negation of description , action and judgment |

We propose 7 semantic relations of language domain which can be managed in our model.

①．We say that event $A$ has inheritance relation with event $B$ if $B$ is a special case of $A$. We denote it $A \xrightarrow{inherit} B$.

②．We say that event $A$ has accompany relation with event $B$ if the occurrence of event $A$ is equal to the occurrence of event $B$. We denote it $A \xrightarrow{accompany} B$.

③．We say that event $A$ has subprocess relation with event $B$ if $B$ is a subevent of $A$ and $B$ occurs in some time segment of $A$. We denote it $A \xrightarrow{part} B$.

④．We say that event $A$ has precondition relation with event $B$ if event $B$ occurs before $A$ occurs. We denote it $A \xrightarrow{pre} B$.

⑤．We say that $B$ has goal relation with event $A$ if the occurrence of $A$ is a motive of event $B$. We denote it $A \xrightarrow{goal} B$.

⑥．We say that $A$ has relation of cause or consequence with event $B$ if the occurrence of $B$ is a cause or consequence of $A$. We denote it $A \xrightarrow{cause} B$ or $A \xrightarrow{cons} B$.

⑦．We say that event $A$ has sequential relation with event $B$ if the event which can occur after event $A$ occurs is always event $B$. We denote it $A \xrightarrow{sequel} B$.

## 2．Memory schema

Let $S_1, S_2, \cdots, S_n$ be $n$ sentences of a text and $E_1, E_2, \cdots, E_n$ events corresponding to it. We suppose that subsequence of events $E_{n1}, E_{n2}, \cdots, E_{nl}$ $(1 \le l \le n)$ of $E_1, E_2, \cdots, E_n$ occurs consequently and denote it by

$$E_{n1} \xrightarrow{sequel} E_{n2} \xrightarrow{sequel} \cdots \xrightarrow{sequel} E_{nl} \quad (*)$$

In our model, understanding of text is a process confirming the semantic relations in (*). This process is performed by matching $l$ events and its relevant events to a story schema, which is defined in our model. We consider this course as a process for text understanding

**Definition 1.** Let $E, \overline{E}$ be two events. We say that $\overline{E}$ matches $E$ if every case-relations and values of $\overline{E}$ are equal to ones of $E$. We call that $\overline{E}$ is event schema and $E$ matchable event.

*Remark:* An event can match another event or can be matched by another event.

Through the matching, the value of case-relation with undefined value in $\overline{E}$ is unified by one in $E$. We call such a unifying process confirmation process of case-relation with undefined value.

**Definition 2.** We call the following relational diagram memory schema and denote by $MP(\overline{E}_1, \overline{E}_2, \cdots \overline{E}_n)$ or $MP$.

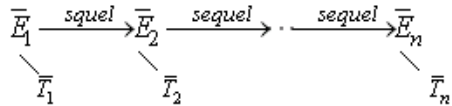

Figure 1．Structure of memory schema

Here, $\overline{T}_i (i = 1, \cdots, n)$ is a tree having $\overline{E}_i$ as its root. Each node of this tree is an event schema or an event. For two neighboring nodes there is a edge with label $r$ denoting semantic relation between them. In addition to $r$, it is possible to add test factor "$"to edge. $\overline{E}_1$ is called a

root of memory schema $MP(\bar{E}_1, \bar{E}_2, \cdots \bar{E}_n)$.

Now, let $\tilde{E}$ be a set of events. Memory mapping $Memory$ is defined as follows:

$$Memory: \quad \tilde{E} \to \{True, False\}$$

We say that $E$ is understandable in memory if $Memory(E) = True$ for $\forall E \in \tilde{E}$. So called "understandable in memory" means that our understanding model senses occurrence (or existence) of this event. We suppose that $Memory$ satisfies the following 3 conditions:

①. $\bar{E}_1, \bar{E}_2$: two event schemas in memory schema $MP$.

If $\bar{E}_1 \xrightarrow{pre,\$} \bar{E}_2$ holds and there exist two event $E_1$, $E_2$ which are matched by $\bar{E}_1$ and $\bar{E}_2$ and $Memory(E_2) = True$, then relational diagram $\bar{E}_1 \xrightarrow{pre} \bar{E}_2$ holds.

②. Let us assume that there exists a following relational diagram in memory schema $MP$.

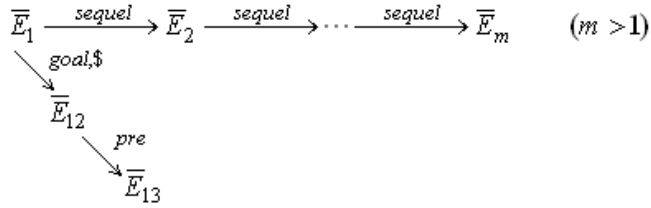

If events $E_1, E_2, \cdots, E_m, E_{12}, E_{13}$ are matched by $\bar{E}_1, \bar{E}_2, \cdots, \bar{E}_m, \bar{E}_{12}, \bar{E}_{13}$,

$Memory(E_1) = Memory(E_2) = \cdots = Memory(E_m) = Memory(E_{13}) = True$

and $Fs(E_m) = E_{13}$, then $Memory(E_{12}) = True$ and the following relational diagram holds.

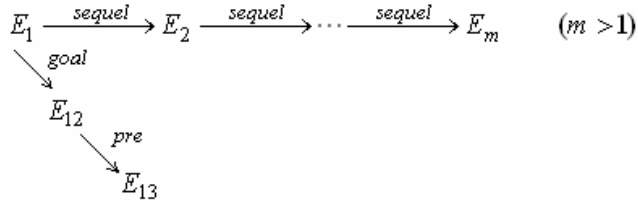

③. Let $\bar{E}_1, \bar{E}_2$ be two event schemas in memory schema $MP$ and $\bar{E}_1 \xrightarrow{r} \bar{E}_2$. Let $E_1, E_2$ be events being matched by $\bar{E}_1, \bar{E}_2$ respectively.

If $Memory(E_1) = True$, then $Memory(E_2) = True$ and relational diagram $E_1 \xrightarrow{r} E_2$ holds.

**Definition 3.** Let $E_1, E_2, \cdots, E_n$ be $n$ events. If $E_1, E_2, \cdots, E_n$ satisfies the following 5 conditions, we call sequence of events constituting of $n$ events is matched by memory schema $MP(\bar{E}_1, \bar{E}_2, \cdots, \bar{E}_k)$.

(1) Each event schema in subsequence $\bar{E}_{k1}, \bar{E}_{k2}, \cdots, \bar{E}_{kl}$ $(1 \leq l \leq k)$ of sequence of events $\bar{E}_1, \bar{E}_2, \cdots, \bar{E}_k$ in $MP(\bar{E}_1, \bar{E}_2, \cdots, \bar{E}_k)$ matches $l$ events in subsequence $E_{n1}, E_{n2}, \cdots, E_{nl}$ of sequence of events $E_1, E_2, \cdots, E_n$ one by one.

(2) Let $Q(E_1, E_2, \cdots, E_n) = \{<E_{ni}> | i = \overline{1,l}\}$ be a quotient set of $E_1, E_2, \cdots, E_n$ where $<E_{ni}>$ is a set of events between $E_{n(i-1)}$ and $E_{n(i+1)}$ and includes $E_{ni}$. That is $<E_{ni}>$ is a set of events of $E_{n(i-1)}$, …, $E_{ni}$, …, $E_{n(i+1)}$ excepting $E_{n(i-1)}$ and $E_{n(i+1)}$.

For $i = 1, 2, \cdots, l$, each event of $<E_{ni}>$ is matched by event schema of tree $\bar{T}_i$ in memory schema $MP(\bar{E}_1, \bar{E}_2, \cdots, \bar{E}_k)$. All remained event schemas unmatched in $\bar{T}_i$ can be event schemas including same variable. We can confirm its values through confirmation process of case-relation with the same variable.

(3) If set $A = \{\bar{E}_1, \bar{E}_2, \cdots, \bar{E}_k\} \setminus \{\bar{E}_{k1}, \bar{E}_{k2}, \cdots, \bar{E}_{kl}\}$ is not empty, event schemas of this set are event schemas including the same variable, We can confirm its values through confirmation process of case-relation with the same variable.

(4) If $E_x \in \{E_1, E_2, \cdots, E_k\}$ is matched by root $\bar{E}_1$ of $MP(\bar{E}_1, \bar{E}_2, \cdots, \bar{E}_k)$, $Memory(\bar{E}_x) = True$ holds.

(5) Let $\bar{E}_{i1}, \bar{E}_{i2}$ be two event schemas of tree $\bar{T}_i$ in $MP(\bar{E}_1, \bar{E}_2, \cdots, \bar{E}_k)$ and we suppose that $\bar{E}_{i1} \xrightarrow{r,\$} \bar{E}_{i2}$, $\bar{E}_{i1}, \bar{E}_{i2}$ can be confirmed through confirmation process of the same variable. And let $E_{i1}, E_{i2} \in <E_{ni}>$ be matched by $\bar{E}_{i1}, \bar{E}_{i2}$ respectively.

If $r = $"pre", then $Memory(E_{i2}) = True$ and if $r = $"goal", then there exists the following relational diagram in memory schema $MP(\bar{E}_1, \bar{E}_2, \cdots, \bar{E}_k)$.

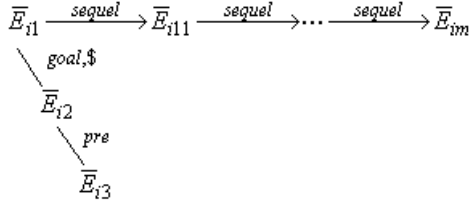

And $FS(\overline{E}_{im}) = \overline{E}_{i3}$ holds.

**Definition 4.** Let $E_1, E_2, \cdots, E_n$ $(n>1)$ be sequence of events. We call that $E_1, E_2, \cdots, E_n$ is understandable in memory if $Memory(E_i) = True$ for all $i(i=\overline{1,n})$ and subsequence $E_{n1}, E_{n2}, \cdots, E_{nl}$ $(1<l<=n)$ of $E_1, E_2, \cdots, E_n$ satisfies the following relational diagram.

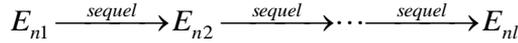

Here, $l$ is not greater than $n$ and must be close to $n$ as much as possible.

**Theorem 1.** Let $E_1, E_2, \cdots, E_n$ $(n>1)$ be sequence of events and $MP$ memory schema matching $E_1, E_2, \cdots, E_n$ $(n>1)$. And let $E_x \in \{E_1, E_2, \cdots, E_n\}$ be event being matched by $\overline{E}_x$ which is a root of $MP$. Then, $E_1, E_2, \cdots, E_n$ is understandable in memory if $E_x$ is understandable in memory

Now, let $E_1, E_2, \cdots, E_n$ be a sequence consisting of $n$ events corresponding to text $P$ and $MP_1, MP_2, \cdots, MP_m$ $m$ memory schemas dividing $E_1, E_2, \cdots, E_n$ into $m$ parts $EP_1, EP_2, \cdots, EP_m$. That is, $MP_i$ matches $EP_i (EP_i = E_{n(i-1)}, \cdots, E_{ni}; i=1,2,\cdots,m; E_{n0} = E_1, E_{nm} = E_n)$. And let root $\overline{E}_{xi}$ of some tree in $MP_i$ and root $\overline{E}_{y(i+1)}$ of some tree in $MP_{i+1}$ satisfy relational diagram $\overline{E}_{xi} \xrightarrow{sequel} \overline{E}_{y(i+1)}$. If $E_{x1}$ is an event matched by root of $MP_1$ corresponding to $EP_1$ and is understandable in memory, then sequence of events $E_1, E_2, \cdots, E_n$ is understandable in memory.

## 3. Story and understanding diagram in memory

Generally, story consists of a series of consequent events and some other events relating to it. To make system understand such a story, we defined memory schema in this paper. That is, a memory schema is a kind of data structure base on this idea. Now, let $E_1, E_2, \cdots, E_n$ be sequence of events and $MP(\bar{E}_1, \bar{E}_2, \cdots, \bar{E}_k)$ be memory schema matching this sequence. Based on the condition (1)~(3) which is given in the definition of matching of memory schema, understanding system can deal with the following cases.

① There exists $E_x, E_x \in \{E_1, E_2, \cdots, E_n\}$ which is matched by $\bar{E}_x$. In this case $\bar{E}_x$ of $MP(\bar{E}_1, \bar{E}_2, \cdots, \bar{E}_k)$ is replaced with $E_x$.

② There exists $E_x, E_x \in \{E_1, E_2, \cdots, E_n\}$ which is matched by $\bar{E}_x$.

Here, $\bar{E}_x$ is an event schema with the same variable and $E_x$ is confirmed through confirmation process of case-relation with the same variable. Also in this case, $\bar{E}_x$ of $MP(\bar{E}_1, \bar{E}_2, \cdots, \bar{E}_k)$ is replaced with $E_x$.

③ $\bar{E}_x$ is an event schema not including case-relation with undefined value. In this case, no one is replaced.

For all event schema of $MP(\bar{E}_1, \bar{E}_2, \cdots, \bar{E}_k)$, we apply the procedure described above and delete test factor "$" of edges in $MP(\bar{E}_1, \bar{E}_2, \cdots, \bar{E}_k)$. We call such a diagram story.

Let us assume that $E_1, E_2, \cdots, E_n$ is a sequence of events in text $P$ and is understandable in memory. Then, we can obtain $m$ stories from $m$ memory schemas by applying consequence 1 of theorem 1. This relational diagram consisting of $m$ stories is called understanding diagram in memory of text $P$ and is denoted by $U(P)$ in our paper.

From the above discussion we can see that understanding process in memory depends on memory schema. Description ability of memory schema depends on the structure of memory schema and semantic relations defined by system(Note that the memory schema of this paper assumes consequent occurrence of events).

## 4. Conclusion

In this paper, we defined event expression over sentences in natural language and semantic relations between events. We also defined event schema, memory schema and understandable sequence of events in memory. Based on this definition, we formally considered text understanding process having understanding diagram in memory.

Script model in [1] assumes some kind of concrete scripts(for example,

meal script and dentist script). These scripts consist of stories of some fixed and multistoried structure. Besides it, scripts assume rich and detailed contents which will be difficult constraints to realize. Memory schema of this paper can be constructed easily compare to [1].

　　We think it is possible to make memory schema if we have a history of some series. Of course, memory schema must be reasonable and standard as much as possible.